\documentclass[11pt]{article}
\usepackage[utf8]{inputenc}
\usepackage[T1]{fontenc}
\usepackage[margin=1in]{geometry}
\usepackage{url,hyperref,microtype,graphicx,booktabs,tabularx,array,float,placeins,amsmath,amssymb}
\usepackage[numbers,sort&compress]{natbib}
\hypersetup{hidelinks}
\title{Image-Domain Poisson-Perturbation Robustness of Ischemic Core/Penumbra--Bearing NCCT Slice Classification: Fixed-Checkpoint and Task-Aware Denoising Evaluations}
\author{%
Rhea Ghosal\textsuperscript{1,*}, Ronok Ghosal\textsuperscript{2}, and Eileen Lou\textsuperscript{3}\\[0.6em]
\small \textsuperscript{1}Westlake High School, Austin, TX, United States\\
\small \textsuperscript{2}California Institute of Technology, Pasadena, CA, United States\\
\small \textsuperscript{3}Round Rock High School, Round Rock, TX, United States\\[0.4em]
\small \textsuperscript{*}Corresponding author: Rhea Ghosal; \href{mailto:grhea1008@gmail.com}{grhea1008@gmail.com}}
\date{}

\begin{document}
\maketitle

\begin{abstract}
Image-domain Poisson perturbation may alter normalized NCCT appearance and downstream models. We evaluated classification of ischemic core/penumbra--bearing non-contrast CT (NCCT) slices under five image-domain simulated settings. The source cohort was CPAISD (112 patients with hyperacute ischemic stroke); CPAISD's predefined test partition contained 10 patients and 809 slices (242 lesion-bearing and 567 mask-negative). First, a fixed-checkpoint audit compared direct ResNet-18 classification (P1) with residual U-Net denoising followed by the same classifier (P2) across noise seeds 7, 17, and 27. P1 average precision (AP) ranged from 0.694 to 0.901, whereas P2 AP ranged from 0.509 to 0.797 and was substantially lower at settings 10--40. The fixed 0.5 threshold had 0--12.8\% sensitivity for P1 and 0\% for P2. Second, a prospectively locked de novo experiment compared direct noisy classification (DNC), joint denoising--classification (JDC-0), and the same joint model with privileged training-only lesion-boundary supervision (JDC-B). All three configurations were trained with five paired seeds and test-evaluated in one complete pass under the prospectively locked protocol. Across-setting mean AP (mean $\pm$ sample SD) was $0.861\pm0.028$ for DNC, $0.840\pm0.038$ for JDC-0, and $0.856\pm0.031$ for JDC-B. Hierarchical paired-bootstrap differences were $-0.021$ (95\% CI $-0.063$ to $0.019$) for JDC-0 minus DNC, $0.016$ ($-0.019$ to $0.054$) for JDC-B minus JDC-0, and $-0.005$ ($-0.041$ to $0.026$) for JDC-B minus DNC; none excluded zero. A frozen stress test on the official 52-patient AISD test partition also showed limited transportability. Thus, ordinary joint training did not demonstrate a classification benefit, and the boundary term recovered part of its point-estimate loss without a statistically supported advantage. Image fidelity, ranking, calibration, and clinical utility must be evaluated separately. This study does not validate acquired low-dose, portable, or cone-beam CT, nor patient-level stroke-versus-control diagnosis.
\end{abstract}

\section{Introduction}

Acquired low-dose CT can reduce radiation exposure, but changes in acquisition and reconstruction may increase noise and alter image appearance. Such changes may affect both human interpretation and downstream artificial-intelligence models. A controlled image-domain perturbation audit can test model sensitivity to altered image statistics, but it is distinct from physical-dose characterization or clinical validation of a scanner or workflow.

We first study two historical fixed-checkpoint inference pathways under image-domain Poisson-perturbed NCCT conditions. Pipeline~1 classifies perturbed slices directly, whereas Pipeline~2 first applies a residual U-Net and then uses the same fixed classifier. This audit isolates the effect of a pre-existing denoiser at inference time. We then test the unresolved training question in a prospectively locked de novo experiment comparing direct perturbed-image classification, end-to-end joint denoising--classification, and the same joint model with a lesion-boundary auxiliary objective.

We focus on a controlled comparison across unitless image-domain simulation settings with matched random seeds and report image quality, classification performance, calibration error, and inference time. The perturbation is image-domain Poisson noise; it does not model scanner-specific noise correlations, reconstruction effects, motion, beam hardening, ring artifacts, or cone-beam CT geometry.

\section{Related Work}

Deep learning has become central to CT imaging, particularly in denoising and pathology detection. Earlier hand-crafted methods such as BM3D \cite{dabov2007image} and Non-Local Means reduce noise but tend to blur subtle anatomical features at low signal-to-noise ratios (SNR).

Models like DnCNN \cite{zhang2017beyond}, RED-Net \cite{mao2016image}, and U-Net \cite{ronneberger2015unet} learn direct mappings from noisy to clean images and report strong Peak Signal-to-Noise Ratio (PSNR) and Structural Similarity Index Measure (SSIM) scores \cite{wang2004ssim}. However, these networks require large datasets and may be difficult to deploy on lightweight or hardware-constrained systems.

Convolutional neural networks such as ResNet \cite{he2016resnet} and DenseNet \cite{huang2017densely} have also been applied to stroke classification from CT and MRI. Most of these studies assume clean, high-quality scans and do not evaluate diagnostic endpoints under controlled image-domain perturbations.

Adjacent literature further motivates the separation of data, reconstruction, and diagnostic questions. Kujur et al. evaluated how data complexity and model dependence interact in brain-MRI classification, reinforcing the need to report cohort composition and model-specific behavior rather than treating a headline accuracy as dataset-independent evidence \cite{kujur2022complexity}. Zubair et al. reviewed deep-learning methods for LDCT denoising and emphasized the continuing challenge of reducing noise and artifacts while preserving diagnostically important structure \cite{zubair2024predication}; this supports our separate reporting of reconstruction fidelity and classification utility. Aung and Khan integrated attention mechanisms into a U-Net for lung-nodule segmentation to improve local and global feature representation \cite{aung2025attention}. The MRI disease-classification and lung-nodule-segmentation studies are methodologically adjacent rather than direct validation of our image-domain-perturbed ischemic-lesion slice endpoint.

Few studies benchmark classification across controlled image-domain noise settings or directly compare a fixed classifier before and after denoising. Here, we evaluate both pathways under controlled image-domain Poisson perturbations. Portable CT scanners are already being used in stroke care, including in ambulance-based stroke units \cite{grotta2014mobile, shuaib2020mobile}. These use cases motivate future external validation; no acquired low-dose, portable-CT, cone-beam-CT, or clinical-workflow data are evaluated in the present study.

Transformer-based models such as SwinIR \cite{liang2021swinir}, Restormer \cite{zamir2022restormer}, and TransUNet \cite{chen2021transunet} extend CNN-based methods by modeling long-range dependencies with self-attention. They achieve strong results in image restoration and biomedical imaging, and motivate task-oriented objectives that jointly optimize denoising and classification under noise and artifact shifts \cite{zhang2021task}.

\section{Materials and Methods}

\subsection{Datasets, Preprocessing, and Endpoint}
The internal cohort was the Core--Penumbra Acute Ischemic Stroke Dataset (CPAISD), comprising 112 anonymized conventional NCCT studies from patients with hyperacute ischemic stroke \cite{umerenkov2025cpaisd}. We retained its predefined patient-disjoint split: 92 patients for training (8,376 slices), 10 for validation (980 slices), and 10 for testing (809 slices). Reference masks encode background (0), ischemic core (1), and penumbra (2). A slice was labeled positive when its mask contained core or penumbra and negative otherwise. Class counts were 1,719/6,657 positive/negative training slices, 222/758 validation slices, and 242/567 test slices. Because all 112 participants had ischemic stroke, a negative observation is a mask-negative slice within a stroke study, not a non-stroke patient. The endpoint is therefore lesion-bearing versus mask-negative slice classification, not hemorrhage detection or patient-level stroke-versus-control diagnosis.

CPAISD DICOM pixel values were converted to Hounsfield units using rescale slope and intercept. Values below 0 HU were clipped, each slice was divided by its positive maximum, and images were evaluated at their native $512\times512$ array resolution. The same frozen normalization was used for both pathways.

\subsection{Robustness Evaluation Protocol}
We evaluated five nominal image-domain simulation settings ($\lambda\in\{1,5,10,20,40\}$) using deterministic seeds 7, 17, and 27. Each seed controlled the Poisson draw for every slice. We report slice-level average precision (AP), ROC--AUC, complete confusion counts and operating metrics, expected calibration error (ECE), PSNR, SSIM, and latency. Primary discrimination analyses averaged the three scores for each slice before metric calculation; seed-wise mean $\pm$ standard deviation is retained for the original performance and reconstruction summaries.

\subsection{Image-Domain Poisson Simulation}
To generate controlled image-domain perturbations, we applied Poisson sampling to each normalized reference NCCT image $I(x,y)$ according to:

\[
I_{\mathrm{sim},\lambda}(x,y) \sim \frac{1}{\lambda}\,\mathrm{Poisson}\!\big(\lambda\, I(x,y)\big)
\]
\noindent Here, $I(x,y)\in[0,1]$ is normalized reference NCCT intensity and $\lambda\in\{1,5,10,20,40\}$ is a unitless simulation index. Under this construction, smaller $\lambda$ produces larger Poisson variance and visibly noisier images. It is not CTDIvol, mAs, a measured physical photon count, or a scanner acquisition setting. Simulated images were generated on the fly, clipped to $[0,1]$, and reproduced with the prespecified random seeds. Dose reduction has been studied in settings such as lung cancer screening, but this image-domain perturbation is not a scanner-physics or projection-domain simulation \cite{mccollough2017lowdose}.

\subsection{Model Architectures}
\begin{figure}[H]
    \centering
    \includegraphics[width=0.95\linewidth]{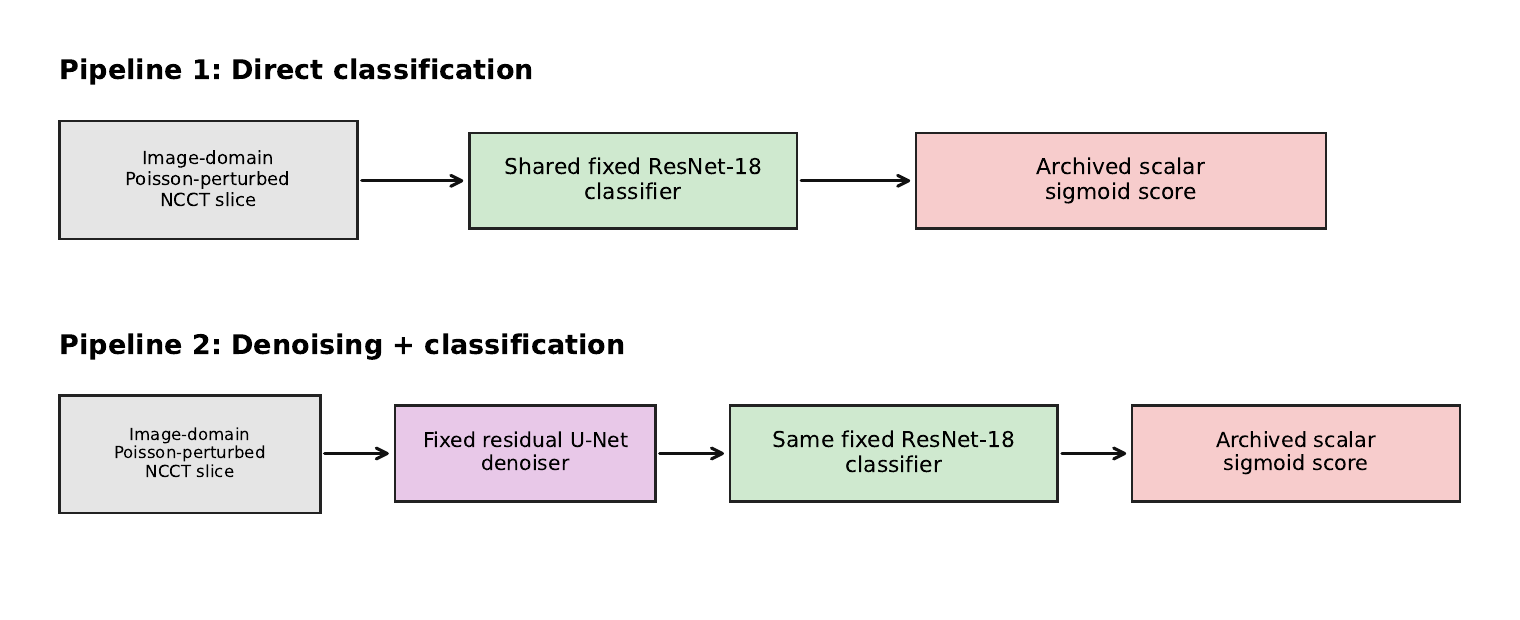}
    \caption{Fixed-checkpoint inference pathways. P1 sends an image-domain Poisson-perturbed NCCT slice directly to the archived ResNet-18 classifier. P2 first applies the archived residual U-Net and then the same ResNet-18 checkpoint. The output is the archived classifier's scalar sigmoid score, evaluated post hoc against CPAISD lesion-bearing versus mask-negative labels. Because the original training target is unavailable, this score is not interpreted as a calibrated ischemic-lesion probability or as patient-level stroke-versus-control classification. The simulation setting is not an acquired dose, portable-CT, or cone-beam-CT measurement.}
    \label{fig:architecture}
\end{figure}

As shown in Figure~\ref{fig:architecture}, we designed two pipelines for comparison:
\textbf{Pipeline~1 (Direct Classification):}  
The archived ResNet-18 takes single-channel image-domain Poisson-perturbed NCCT images as input and produces a scalar sigmoid score. Because its pre-archive training dataset and target-label definition are unavailable, we evaluate this score against the CPAISD lesion-bearing versus mask-negative labels solely as a descriptive fixed-checkpoint audit; it must not be interpreted as a calibrated probability of ischemic-lesion presence. ResNet-18 was selected in the original implementation as the classifier architecture. 
\textbf{Pipeline~2 (Denoising + Classification):}  
Before classification, we add a denoising step. The denoiser is a residual U-Net, an encoder--decoder network with skip connections whose bounded residual is added to the noisy input. Both pathways use the identical archived ResNet-18 checkpoint and the same $[-1,1]$ classifier normalization. Pipeline~2 therefore has no separately initialized classifier, and no P2-specific classifier fine-tuning or joint denoiser--classifier training was performed for this fixed-checkpoint audit. The comparison isolates the effect of preprocessing at inference time rather than retraining the classifier for denoised inputs. The pre-archive training history of the shared classifier and denoiser is unavailable and was not retrospectively reconstructed. Their pre-archive training-data provenance, including dataset identity and patient identifiers, is also unavailable. Consequently, although CPAISD's released partitions are mutually patient-disjoint, we cannot verify that the archived checkpoints were unexposed to CPAISD validation or test patients; fixed-checkpoint P1/P2 results are therefore descriptive audits rather than confirmed leakage-free out-of-sample estimates.

The historical fixed-checkpoint audit and the de novo experiment have different evidentiary roles. The historical documentation described Adam optimization with a learning rate of $10^{-4}$, batch size 32, at most 50 epochs, and early stopping on validation loss. However, the evaluated classifier and denoiser checkpoints contain weights only; the retained artifacts do not preserve either loss definition, classifier class weighting or sampling, optimizer state or history, patience or minimum-improvement rule, or complete checkpoint-selection history. In particular, we cannot verify whether the historical classifier used weighted BCE or whether the historical denoiser used pure $L_2$/MSE, SSIM, perceptual, or another reconstruction objective, and therefore do not attribute its failure to a particular classifier loss or weighting scheme or to $L_2$-induced oversmoothing. We treat the original settings as an unverified historical description rather than a reproducible execution record. The losses, coefficients, optimizer, checkpoint criterion, and stopping rule reported below apply only to the de novo DNC/JDC experiments; they are not retroactively attributed to the archived P1/P2 weights. We interpret P1 versus P2 only as an inference-stage component audit, whereas conclusions about joint training and the boundary objective derive only from the fully logged de novo runs.

Both models were implemented in PyTorch and evaluated on the native $512\times512$ grayscale arrays without interpolation.

\subsection{De novo Task-Aware Denoising--Classification Experiment}
\label{sec:task_aware_experiment}

We performed a prospectively specified, de novo experiment to distinguish end-to-end denoising--classification training from the incremental effect of a lesion-boundary auxiliary loss. The endpoint and patient-disjoint CPAISD partitions were unchanged. During training, one setting in $\{1,5,10,20,40\}$ was sampled uniformly for each slice and epoch using a reproducible generator keyed by training seed, epoch, and sorted slice index; no additional geometric or intensity augmentation was used. Validation and test inference evaluated all five settings with noise seeds 7, 17, and 27, averaging the three probabilities by slice before metric calculation. All configurations operated on native $512\times512$ arrays. The simulation settings are image-domain indices, not CTDIvol, mAs, or physical acquisition-dose measurements.

We compared three matched configurations (Table~\ref{tab:task_aware_configurations}). Direct noisy classification (DNC) passed the simulated input directly to a ResNet-18 classifier and minimized unweighted binary cross-entropy with logits, $L_{\mathrm{BCE}}$. Joint denoising--classification without a boundary term (JDC-0) passed the input through a residual U-Net and then to the classifier. For slice $i$, the support-normalized reconstruction loss was
\begin{equation}
\ell_{\mathrm{rec},i}=\frac{\sum A_i\lvert\hat{x}_i-x_i\rvert}{\sum A_i+10^{-8}},
\qquad A_i=\mathbb{1}[x_i>0],
\qquad L_{\mathrm{rec}}=\frac{1}{N}\sum_{i=1}^{N}\ell_{\mathrm{rec},i},
\end{equation}
and $L_{\mathrm{JDC0}}=L_{\mathrm{BCE}}+\lambda_{\mathrm{rec}}L_{\mathrm{rec}}$.

JDC-B was architecturally identical to JDC-0 but added privileged, training-only lesion-boundary supervision. The lesion mask $M=\mathbb{1}[\mathrm{mask}>0]$ was used only to construct a boundary band $B=\operatorname{dilate}_{5\times5}(M)-\operatorname{erode}_{5\times5}(M)$; it was never supplied as a model input and was unavailable at inference. With fixed $3\times3$ Sobel operators $G_x$ and $G_y$, the per-slice boundary loss was
\begin{equation}
\ell_{\mathrm{bnd},i}=\frac{\sum B_i\left(\lvert G_x(\hat{x}_i)-G_x(x_i)\rvert+\lvert G_y(\hat{x}_i)-G_y(x_i)\rvert\right)}{2\sum B_i+10^{-8}},
\qquad L_{\mathrm{bnd}}=\frac{1}{N}\sum_{i=1}^{N}\ell_{\mathrm{bnd},i}.
\end{equation}
A slice without boundary pixels contributed a differentiable zero. Reconstruction and Sobel calculations were performed per slice in FP32 outside automatic mixed precision. The complete objective was $L_{\mathrm{JDCB}}=L_{\mathrm{BCE}}+\lambda_{\mathrm{rec}}L_{\mathrm{rec}}+\lambda_{\mathrm{bnd}}L_{\mathrm{bnd}}$. Thus, JDC-0 versus DNC tested end-to-end task-aware training, whereas JDC-B versus JDC-0 isolated this specific boundary term.

The classifier used torchvision ResNet-18 initialized with verified ImageNet-1K V1 weights. Its three-channel first-layer kernels were averaged to initialize a one-channel convolution, and its binary head was newly initialized for every training seed. The complete classifier was trainable in all configurations. The complete U-Net was trainable in JDC-0 and JDC-B, with its output convolution initialized so that the residual path began as the identity. Within each seed, configurations shared the same classifier initialization, and JDC-0 and JDC-B also shared the same U-Net initialization. Historical checkpoints were not used for initialization.

Only auxiliary-loss coefficients were selected in a validation-only pilot using seed 991, which was not a reported training seed. We compared $\lambda_{\mathrm{rec}}\in\{0.1,1.0\}$ and then $\lambda_{\mathrm{bnd}}\in\{0.05,0.20\}$, selecting the larger validation across-setting mean AP; differences below 0.005 were resolved in favor of the smaller coefficient. The selected values were $\lambda_{\mathrm{rec}}=0.1$ and $\lambda_{\mathrm{bnd}}=0.20$. Pilot checkpoints were quarantined as test-ineligible and were never evaluated on test.

Final runs used AdamW, learning rates of $10^{-4}$ for the classifier backbone and U-Net and $3\times10^{-4}$ for the new head, weight decay $10^{-4}$, micro-batches of 4 with four-step accumulation (effective batch size 16), automatic mixed precision for classification, global gradient clipping at 1.0, and cosine decay to $10^{-6}$ over at most 40 epochs. The checkpoint maximizing validation across-setting mean AP was retained. Early stopping was permitted after 10 epochs and followed eight consecutive epochs without at least 0.001 improvement. We completed all three configurations for five paired seeds (101, 211, 307, 401, and 503); the prespecified three-seed fallback was not used.

The primary estimand was across-setting mean test AP, defined as the unweighted mean of AP across the five unitless image-domain simulation settings. Percentile 95\% confidence intervals for paired contrasts used 2,000 hierarchical paired-bootstrap replicates (random seed 20260904), resampling both training seeds and the 10 test patients while retaining all slices from each sampled patient. Secondary metrics included setting-specific AP and ROC--AUC, Brier score, 15-bin equal-width ECE, and confusion-matrix metrics at 0.5 and at the highest validation-only threshold retaining at least 90\% sensitivity. Reconstruction metrics for JDC-0 and JDC-B included PSNR, SSIM, whole-support MAE, and boundary-gradient MAE.

\begin{table}[!htbp]
\centering
\caption{Locked de novo task-aware configurations. BCE was unweighted in every arm. Auxiliary losses were normalized per slice and averaged over the micro-batch in FP32. JDC-B used lesion masks as privileged training-only supervision; masks were neither model inputs nor available during inference.}
\label{tab:task_aware_configurations}
{\footnotesize
\setlength{\tabcolsep}{3pt}
\renewcommand{\arraystretch}{1.12}

\textit{Architecture and initialization}\par\smallskip
\begin{tabularx}{\textwidth}{@{}l
  >{\raggedright\arraybackslash}X
  >{\raggedright\arraybackslash}X
  >{\raggedright\arraybackslash}X@{}}
\toprule
Configuration & Image path & Initialization & Trainable modules \\
\midrule
DNC & Simulated image $\rightarrow$ ResNet-18
    & Paired ImageNet-1K V1 classifier; new binary head
    & Complete classifier \\
JDC-0 & Image $\rightarrow$ residual U-Net $\rightarrow$ ResNet-18
      & Same classifier; paired identity-initialized U-Net
      & Complete U-Net and classifier \\
JDC-B & Same as JDC-0
      & Same paired classifier and U-Net as JDC-0
      & Complete U-Net and classifier \\
\bottomrule
\end{tabularx}

\medskip
\textit{Objective and checkpoint selection}\par\smallskip
\begin{tabularx}{\textwidth}{@{}l
  >{\raggedright\arraybackslash}p{0.28\textwidth}
  >{\raggedright\arraybackslash}X@{}}
\toprule
Configuration & Objective & Validation checkpoint/stopping \\
\midrule
DNC & $L_{\mathrm{BCE}}$
    & Best validation across-setting mean AP; $\geq10$ epochs; patience 8; minimum $\Delta=0.001$ \\
JDC-0 & $L_{\mathrm{BCE}}+0.1L_{\mathrm{rec}}$
      & Best validation across-setting mean AP; $\geq10$ epochs; patience 8; minimum $\Delta=0.001$ \\
JDC-B & $L_{\mathrm{BCE}}+0.1L_{\mathrm{rec}}+0.20L_{\mathrm{bnd}}$
      & Best validation across-setting mean AP; $\geq10$ epochs; patience 8; minimum $\Delta=0.001$ \\
\bottomrule
\end{tabularx}
}
\end{table}

\subsection{Reproducibility}
The fixed-checkpoint audit artifact records SHA-256 hashes for the historical classifier and denoiser, downloaded AISD source archives, official external test identifiers, raw predictions, generated tables, dataset-split inventory, software/hardware environment, noise seeds, simulation settings, label mappings, and threshold provenance. The original fixed-checkpoint training logs remain unavailable and were not reconstructed. Separately, the de novo task-aware experiment retained immutable split manifests, source and dependency hashes, paired initialization hashes, epoch-level losses and validation metrics, random-number-generator states, checkpoint-selection records, and final checkpoint hashes for all 15 runs. Its auxiliary weights were selected only in a test-ineligible validation pilot; the five-seed training matrix and thresholds were locked before a single complete test evaluation on one full NVIDIA A100 GPU. Legacy checkpoints were provenance records only and did not initialize the new models. The de novo experiment used no test labels for model, hyperparameter, checkpoint, or threshold selection. In the current fixed-checkpoint audit, CPAISD test labels were used only for final evaluation; however, prior exposure during the checkpoints' undocumented pre-archive training cannot be excluded.

Generative AI technology (OpenAI GPT-5, accessed through Codex) assisted with drafting and reviewing analysis code, reproducibility checks, and language and LaTeX editing. The authors executed the analyses, checked source data and machine-readable outputs, verified the numerical claims against sealed artifacts, reviewed every AI-assisted change, and take responsibility for the final methods, results, citations, and figures.

\subsection{Experimental Setup}
We loaded the fixed ResNet-18 classifier and residual U-Net once and evaluated two prespecified pathways: P1, direct classification of image-domain Poisson-perturbed NCCT slices; and P2, residual U-Net denoising followed by the same classifier. For each internal simulation-setting/noise-seed combination, both pathways received identical inputs. The available artifact does not establish the historical classifier loss, optimizer, class weighting, sampling strategy, augmentation, early-stopping rule, or checkpoint-selection rule; this limits causal interpretation of the setting-dependent response.

For an independent conventional-NCCT stress test, we used the authors' official 52-patient test partition of the Acute Ischemic Stroke Dataset (AISD) \cite{liang2021symmetry}. We downloaded the original NCCT DICOM archives, converted pixels to HU using DICOM rescale parameters, and then applied the same frozen preprocessing used for CPAISD. A preliminary screen using the authors' pre-windowed PNG derivatives was excluded after its intensity distribution was found incompatible with the DICOM/HU pipeline. No AISD image or label was used for training, model selection, preprocessing changes, calibration, or threshold selection. The primary AISD endpoint treated mask labels \{2,3,5\} as visible acute-infarct positive (504 positive and 1,002 mask-negative slices); a prespecified sensitivity analysis used \{1,2,3,5\} (553 positive and 953 mask-negative slices). We evaluated unperturbed reference NCCT and the same five perturbation settings and seeds through P1 and P2. AISD and CPAISD labels are clinically related but not identical; AISD does not separate core from penumbra.

\subsection{Evaluation Metrics}
\textbf{Denoising.} We compared P2 output with each reference using PSNR and SSIM \cite{wang2004ssim}; latency was measured per slice.

\textbf{Discrimination.} Because lesion-positive and mask-negative slices were imbalanced, we report average precision (AP), the non-interpolated precision--recall summary, together with ROC--AUC. For primary image-domain simulation analyses, scores were first averaged across seeds 7, 17, and 27 within each slice. AP and ROC--AUC are ranking measures and do not imply a usable operating point.

\textbf{Operating points and calibration.} We report $TP$, $FP$, $TN$, and $FN$, sensitivity, and specificity at the prespecified 0.5 threshold. The confusion counts permit derivation of other fixed-threshold metrics when defined; precision is undefined for all-negative predictions unless a zero-division convention is imposed. Separately, for each pathway and simulation setting, we selected on CPAISD validation the highest threshold retaining at least 90\% sensitivity and applied it unchanged to CPAISD test and AISD. No test or AISD label was used to tune a threshold. ECE and Brier scores were used as descriptive calibration summaries.

\textbf{Uncertainty.} Percentile 95\% confidence intervals used 2,000 patient-cluster bootstrap replicates: patients were sampled with replacement and all slices from each selected patient were retained. Internal AP intervals and external AP/ROC--AUC intervals were calculated after seed averaging. External P2-minus-P1 differences were calculated within the same resamples. This preserves within-patient dependence but cannot compensate for small numbers of independent patients. Patient-level stroke-versus-control metrics are undefined because neither evaluation cohort contains suitable non-stroke control patients.

\textbf{Post-hoc Grad-CAM audit.} We generated gradient-weighted class-activation maps (Grad-CAM) from the final convolutional layer of the shared ResNet-18 classifier using the positive-class logit \cite{selvaraju2017gradcam}. Channel weights were the spatial means of the corresponding gradients; the weighted activation maps were summed, rectified, bilinearly resized to $512\times512$, and normalized within each slice. The cohort-wide audit used the exact canonical seed-7 input for each of all 242 lesion-bearing test slices at every simulation setting. A pointing hit placed the maximum activation inside the core--penumbra union; attribution mass was the normalized nonnegative map mass inside that mask; top-10\% IoU and Dice compared the lesion mask with pixels at or above the within-map 90th percentile; and peak--centroid distance was divided by the image diagonal. These post-hoc measures were not used for training or selection and are descriptive attribution checks rather than segmentation, validated localization, or causal tests. Attribution stability across noise seeds was not evaluated.

\section{Results}
\label{sec:results}

\subsection{Internal Cohort and Slice-Level Discrimination}
The patient-disjoint training, validation, and test partitions contained 1,719/6,657, 222/758, and 242/567 lesion-positive/mask-negative slices, respectively. Test positive prevalence was 29.9\%, which is the no-skill AP reference. All 10 test patients contributed at least one lesion-bearing slice, so patient-level stroke-versus-control discrimination could not be estimated.

\begin{figure}[H]
    \centering
    \includegraphics[width=0.82\linewidth]{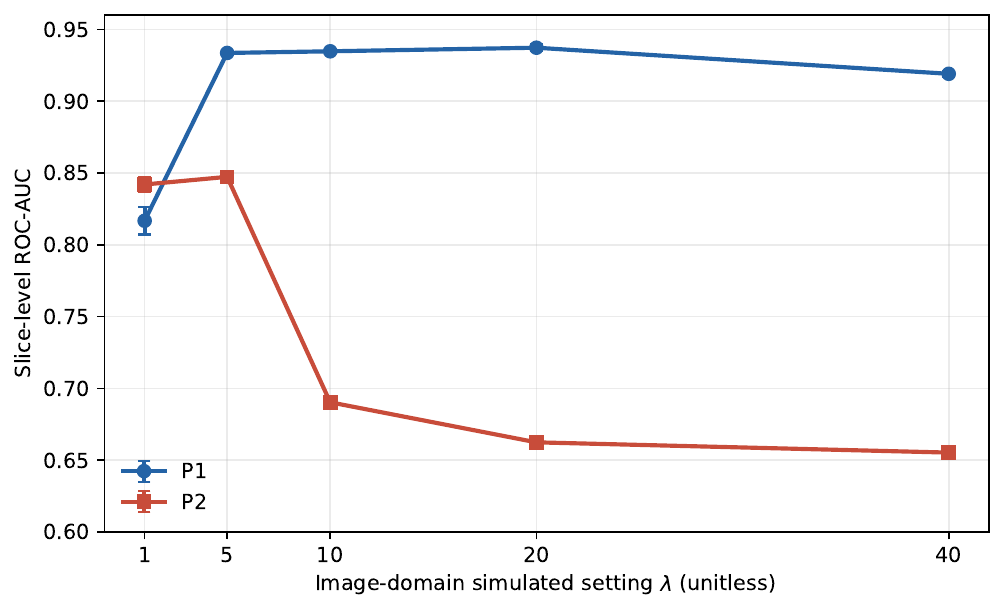}
    \caption{Descriptive seed-wise slice-level ROC--AUC for the fixed-checkpoint pathways across unitless image-domain Poisson simulation settings. Points are the mean of three evaluation-noise-seed-specific ROC--AUC values; error bars are their sample SD. These error bars reflect noise-realization variation only, not patient sampling or model-training variability. Primary ROC--AUC point estimates after within-slice score averaging are reported in Table~\ref{tab:internal_discrimination}.}
    \label{fig:performance}
\end{figure}

After averaging predictions across seeds, P1 achieved AP values of 0.6940, 0.8868, 0.8965, 0.9005, and 0.8805 at settings 1, 5, 10, 20, and 40; corresponding ROC--AUC point estimates were 0.8279, 0.9366, 0.9375, 0.9393, and 0.9224. P2 achieved AP values of 0.7878, 0.7965, 0.5825, 0.5196, and 0.5092, with ROC--AUC values of 0.8484, 0.8527, 0.6906, 0.6621, and 0.6553. Thus, P2 ranked slices slightly better only at setting 1 and was substantially worse at settings 10--40 (Table~\ref{tab:internal_discrimination}). Figure~\ref{fig:performance} separately shows the mean and sample SD of the three noise-seed-specific ROC--AUC values; these descriptive seed error bars are not confidence intervals. These are slice-ranking results within an all-stroke cohort, not patient-level diagnostic performance.

\begin{table}[H]
\caption{CPAISD slice-level discrimination after averaging seeds 7, 17, and 27. AP intervals use 2,000 patient-cluster bootstrap replicates; ROC--AUC entries are point estimates.}
\label{tab:internal_discrimination}
\centering
\footnotesize
\begin{tabular}{ccccc}
\toprule
Setting & Path & AP & AP 95\% CI & ROC--AUC \\
\midrule
1  & P1 & 0.6940 & 0.5389--0.8302 & 0.8279 \\
1  & P2 & 0.7878 & 0.6158--0.9086 & 0.8484 \\
5  & P1 & 0.8868 & 0.7737--0.9544 & 0.9366 \\
5  & P2 & 0.7965 & 0.6484--0.8912 & 0.8527 \\
10 & P1 & 0.8965 & 0.8111--0.9533 & 0.9375 \\
10 & P2 & 0.5825 & 0.3726--0.7354 & 0.6906 \\
20 & P1 & 0.9005 & 0.8215--0.9548 & 0.9393 \\
20 & P2 & 0.5196 & 0.3220--0.6730 & 0.6621 \\
40 & P1 & 0.8805 & 0.7865--0.9457 & 0.9224 \\
40 & P2 & 0.5092 & 0.3096--0.6564 & 0.6553 \\
\bottomrule
\end{tabular}
\end{table}

\subsection{Internal Operating-Point Audit}
The prespecified 0.5 threshold was not usable. P1 produced $TP/FP/TN/FN$ counts of 0/0/567/242, 31/0/567/211, 17/0/567/225, 8/0/567/234, and 0/0/567/242 at settings 1, 5, 10, 20, and 40, corresponding to sensitivities of 0\%, 12.8\%, 7.0\%, 3.3\%, and 0\%. P2 predicted every slice negative at all settings (0/0/567/242). For P2's 242 lesion-bearing slices, the median three-seed-averaged scores were 0.001029, 0.000112, 0.0000161, 0.0000148, and 0.0000143 at settings 1, 5, 10, 20, and 40, respectively; the largest P2 score across any of these conditions was 0.189031. Thus, every P2 score remained below 0.5. Specificity was 100\% only because neither pathway produced false positives. Useful ranking at some settings therefore coexisted with severe score compression and threshold failure. Full class-conditional score summaries are supplied as Supplementary Data~S4 (\nolinkurl{fixed_checkpoint_score_distributions.csv}).

CPAISD-validation-locked thresholds were extremely small. When applied unchanged to test, P1 sensitivity was 97.9--100\% but specificity only 4.9--37.4\%; P2 sensitivity was 100\% with specificity only 0.5--1.1\% (Table~\ref{tab:internal_thresholds}). High sensitivity was recovered only by classifying most mask-negative slices as positive; this shows severe lack of score alignment and calibration relative to the CPAISD audit endpoint, not a clinical remedy.

\begin{table}[!t]
\caption{CPAISD operating-point audit. Each locked threshold was selected only on validation to retain at least 90\% sensitivity and then applied unchanged to test.}
\label{tab:internal_thresholds}
\centering
\scriptsize
\begin{tabular}{cccccc}
\toprule
Setting & Path & $TP/FP/TN/FN$ at 0.5 & Locked threshold & Sensitivity & Specificity \\
\midrule
1  & P1 & 0/0/567/242  & $1.3301\times10^{-5}$ & 0.9793 & 0.2945 \\
1  & P2 & 0/0/567/242  & $5.4215\times10^{-7}$ & 1.0000 & 0.0053 \\
5  & P1 & 31/0/567/211 & $5.8998\times10^{-6}$ & 0.9835 & 0.3739 \\
5  & P2 & 0/0/567/242  & $4.2275\times10^{-7}$ & 1.0000 & 0.0053 \\
10 & P1 & 17/0/567/225 & $1.5262\times10^{-6}$ & 1.0000 & 0.1146 \\
10 & P2 & 0/0/567/242  & $8.3222\times10^{-8}$ & 1.0000 & 0.0106 \\
20 & P1 & 8/0/567/234  & $9.3920\times10^{-7}$ & 1.0000 & 0.0794 \\
20 & P2 & 0/0/567/242  & $9.6345\times10^{-8}$ & 1.0000 & 0.0106 \\
40 & P1 & 0/0/567/242  & $6.7237\times10^{-7}$ & 1.0000 & 0.0494 \\
40 & P2 & 0/0/567/242  & $9.1016\times10^{-8}$ & 1.0000 & 0.0088 \\
\bottomrule
\end{tabular}
\end{table}

\subsection{Task-Aware Experiment Results}
\label{sec:task_aware_results}

We completed five paired training seeds for each of DNC, JDC-0, and JDC-B. The validation-only pilot selected $\lambda_{\mathrm{rec}}=0.1$ and $\lambda_{\mathrm{bnd}}=0.20$; all pilot checkpoints were quarantined as test-ineligible before final test inference. All configurations, training seeds, and simulated settings are reported, including unfavorable and null results. Primary across-setting estimates and paired contrasts are summarized in Table~\ref{tab:task_aware_primary}.

\begin{table}[!htbp]
\centering
\caption{Primary de novo task-aware results on the patient-disjoint CPAISD test partition. Across-setting mean AP is the unweighted mean across settings 1, 5, 10, 20, and 40 after averaging predictions over evaluation-noise seeds 7, 17, and 27 by slice. Configuration rows show mean $\pm$ sample SD across five paired training seeds and a patient-cluster bootstrap interval. Contrast intervals are percentile 95\% intervals from 2,000 hierarchical paired-bootstrap replicates resampling training seeds and patients.}
\label{tab:task_aware_primary}
\resizebox{\linewidth}{!}{%
\begin{tabular}{lccc}
\toprule
Configuration or paired contrast & Paired seeds, $n$ & Across-setting mean AP, mean $\pm$ SD & Estimate or difference (95\% CI) \\
\midrule
DNC & 5 & $0.8606 \pm 0.0284$ & $0.8606$ ($0.7618$, $0.9257$) \\
JDC-0 & 5 & $0.8399 \pm 0.0378$ & $0.8399$ ($0.7333$, $0.9206$) \\
JDC-B & 5 & $0.8557 \pm 0.0308$ & $0.8557$ ($0.7562$, $0.9232$) \\
\midrule
JDC-0 $-$ DNC (primary) & 5 & --- & $-0.0207$ ($-0.0625$, $0.0188$) \\
JDC-B $-$ JDC-0 (key secondary) & 5 & --- & $0.0158$ ($-0.0188$, $0.0542$) \\
JDC-B $-$ DNC (full system) & 5 & --- & $-0.0049$ ($-0.0414$, $0.0265$) \\
\bottomrule
\end{tabular}
}
\end{table}

None of the three prespecified paired-comparison intervals excluded zero. The primary JDC-0-minus-DNC result therefore did not support an end-to-end classification benefit. Likewise, this experiment did not support either an incremental benefit from the tested boundary objective or a full-system JDC-B benefit over DNC. DNC had the highest point estimate across seeds, although individual-seed ordering varied (Table~\ref{tab:task_aware_per_seed}).

\begin{table}[t]
\centering
\caption{Across-setting mean AP for each independently trained seed on the locked CPAISD test partition.}
\label{tab:task_aware_per_seed}
\begin{tabular}{lccc}
\toprule
Training seed & DNC & JDC-0 & JDC-B \\
\midrule
101 & 0.8718 & 0.8713 & 0.8657 \\
211 & 0.8594 & 0.8115 & 0.8416 \\
307 & 0.8758 & 0.8885 & 0.8894 \\
401 & 0.8837 & 0.8246 & 0.8720 \\
503 & 0.8123 & 0.8033 & 0.8099 \\
\midrule
Mean $\pm$ SD & $0.8606 \pm 0.0284$ & $0.8399 \pm 0.0378$ & $0.8557 \pm 0.0308$ \\
\bottomrule
\end{tabular}
\end{table}

Per-setting AP and ROC--AUC were stable within each configuration (Table~\ref{tab:task_aware_dose}). At the fixed 0.5 threshold, the ranges of the five per-setting mean sensitivities/specificities were 0.664--0.687/0.932--0.951 for DNC, 0.678--0.716/0.910--0.936 for JDC-0, and 0.632--0.674/0.927--0.948 for JDC-B. At thresholds selected only on validation to retain at least 90\% sensitivity, the held-out test sensitivity/specificity ranges were 0.948--0.955/0.673--0.680, 0.935--0.950/0.615--0.632, and 0.939--0.949/0.690--0.726, respectively. These operating points were evaluated without test-set recalibration and are descriptive rather than clinically validated.

\begin{table}[!htbp]
\centering
\caption{Slice-level AP and ROC--AUC by simulated image-domain setting on the locked CPAISD test partition. Values are mean $\pm$ sample SD across five independent training seeds after averaging the three fixed evaluation-noise realizations by slice. Setting numbers are simulation indices, not physical acquisition-dose measurements.}
\label{tab:task_aware_dose}
\resizebox{\textwidth}{!}{%
\begin{tabular}{lcccccc}
\toprule
& \multicolumn{2}{c}{DNC} & \multicolumn{2}{c}{JDC-0} & \multicolumn{2}{c}{JDC-B} \\
\cmidrule(lr){2-3}\cmidrule(lr){4-5}\cmidrule(lr){6-7}
Simulated setting & AP & ROC--AUC & AP & ROC--AUC & AP & ROC--AUC \\
\midrule
1  & $0.8614 \pm 0.0238$ & $0.9286 \pm 0.0212$ & $0.8380 \pm 0.0328$ & $0.9167 \pm 0.0242$ & $0.8552 \pm 0.0252$ & $0.9290 \pm 0.0196$ \\
5  & $0.8625 \pm 0.0304$ & $0.9281 \pm 0.0245$ & $0.8395 \pm 0.0407$ & $0.9177 \pm 0.0244$ & $0.8576 \pm 0.0319$ & $0.9274 \pm 0.0232$ \\
10 & $0.8621 \pm 0.0302$ & $0.9271 \pm 0.0260$ & $0.8407 \pm 0.0398$ & $0.9174 \pm 0.0237$ & $0.8577 \pm 0.0335$ & $0.9275 \pm 0.0239$ \\
20 & $0.8595 \pm 0.0320$ & $0.9244 \pm 0.0282$ & $0.8407 \pm 0.0406$ & $0.9169 \pm 0.0233$ & $0.8558 \pm 0.0339$ & $0.9273 \pm 0.0237$ \\
40 & $0.8575 \pm 0.0280$ & $0.9232 \pm 0.0276$ & $0.8404 \pm 0.0407$ & $0.9170 \pm 0.0223$ & $0.8522 \pm 0.0315$ & $0.9276 \pm 0.0225$ \\
\bottomrule
\end{tabular}}
\end{table}

The two joint configurations had similar reconstruction errors at every simulated setting (Table~\ref{tab:task_aware_reconstruction}). JDC-B did not consistently improve PSNR, SSIM, whole-support MAE, or boundary-gradient MAE over JDC-0. These descriptive image-quality summaries do not replace the paired classification contrast and do not identify the cause of the historical fixed-checkpoint sensitivity failure.

\begin{table}[!htbp]
\centering
\caption{Reconstruction metrics for the two joint configurations on the locked CPAISD test partition. Values are mean $\pm$ sample SD across five training seeds; within each seed, each simulated setting summarizes 809 slices over three fixed evaluation-noise realizations. Lower is better for whole-support and boundary-gradient MAE; higher is better for PSNR and SSIM. No significance claim is based on this descriptive table.}
\label{tab:task_aware_reconstruction}
\scriptsize
\resizebox{\textwidth}{!}{%
\begin{tabular}{llcccc}
\toprule
Setting & Configuration & PSNR (dB) & SSIM & Whole-support MAE & Boundary-gradient MAE \\
\midrule
1  & JDC-0 & $18.1004 \pm 0.5051$ & $0.3490 \pm 0.0605$ & $0.12493 \pm 0.01031$ & $0.03163 \pm 0.00093$ \\
1  & JDC-B & $17.9276 \pm 0.5995$ & $0.3242 \pm 0.0213$ & $0.12812 \pm 0.01282$ & $0.03175 \pm 0.00098$ \\
5  & JDC-0 & $22.6742 \pm 0.1470$ & $0.3559 \pm 0.0628$ & $0.08035 \pm 0.00104$ & $0.02278 \pm 0.00006$ \\
5  & JDC-B & $22.6534 \pm 0.1618$ & $0.3312 \pm 0.0224$ & $0.08024 \pm 0.00156$ & $0.02273 \pm 0.00008$ \\
10 & JDC-0 & $24.9028 \pm 0.0546$ & $0.3876 \pm 0.0639$ & $0.06413 \pm 0.00063$ & $0.01804 \pm 0.00007$ \\
10 & JDC-B & $24.8774 \pm 0.1390$ & $0.3626 \pm 0.0229$ & $0.06403 \pm 0.00089$ & $0.01800 \pm 0.00003$ \\
20 & JDC-0 & $27.1734 \pm 0.1561$ & $0.4280 \pm 0.0651$ & $0.05048 \pm 0.00150$ & $0.01358 \pm 0.00011$ \\
20 & JDC-B & $27.0999 \pm 0.3219$ & $0.4025 \pm 0.0233$ & $0.05065 \pm 0.00190$ & $0.01356 \pm 0.00010$ \\
40 & JDC-0 & $29.3572 \pm 0.3727$ & $0.4682 \pm 0.0663$ & $0.03948 \pm 0.00220$ & $0.00984 \pm 0.00016$ \\
40 & JDC-B & $29.2138 \pm 0.6083$ & $0.4423 \pm 0.0237$ & $0.03997 \pm 0.00303$ & $0.00983 \pm 0.00016$ \\
\bottomrule
\end{tabular}}
\end{table}

Complete per-seed and per-setting discrimination, calibration, confusion, and operating-point metrics are provided in Supplementary Data~S1 (\nolinkurl{per_seed_condition_metrics.csv}); the complete per-seed JDC-0 and JDC-B reconstruction outputs, including the unperturbed reference condition, are provided in Supplementary Data~S2 (\nolinkurl{per_seed_reconstruction_metrics.csv}).

\subsection{External AISD Stress Test}
The official AISD test partition contained 52 patients and 1,506 slices, of which 504 were positive under the primary \{2,3,5\} mapping. On unperturbed DICOM/HU NCCT, both pathways were near chance (P1 AP 0.3205 and ROC--AUC 0.4676; P2 AP 0.3078 and ROC--AUC 0.4524). Under the five simulated perturbations, P1 AP ranged from 0.4922 to 0.5826 and ROC--AUC from 0.6383 to 0.7504 (Table~\ref{tab:external_discrimination}). The counterintuitive difference from reference NCCT must not be interpreted as a benefit of dose reduction; it demonstrates sensitivity to the input distribution.

P2 provided no consistent external benefit. At setting 1, paired P2-minus-P1 differences were small and inconclusive for AP ($0.0235$, 95\% CI $-0.0221$ to 0.0711) and ROC--AUC ($0.0244$, $-0.0085$ to 0.0596). At settings 5--40, P2 was significantly worse: paired AP differences ranged from $-0.2464$ to $-0.1132$ and paired ROC--AUC differences from $-0.2457$ to $-0.1344$; every corresponding 95\% interval excluded zero.

\begin{table}[!htbp]
\caption{External AISD discrimination using original DICOM/HU NCCT and the primary visible-acute label mapping \{2,3,5\}. Intervals use 2,000 patient-cluster bootstrap replicates. Reference denotes unperturbed NCCT.}
\label{tab:external_discrimination}
\centering
\scriptsize
\resizebox{\linewidth}{!}{%
\begin{tabular}{ccccc}
\toprule
Input & P1 AP (95\% CI) & P2 AP (95\% CI) & P1 ROC--AUC (95\% CI) & P2 ROC--AUC (95\% CI) \\
\midrule
Reference & 0.3205 (0.2755--0.3713) & 0.3078 (0.2645--0.3559) & 0.4676 (0.3943--0.5353) & 0.4524 (0.3819--0.5244) \\
1  & 0.4922 (0.3941--0.5979) & 0.5157 (0.4277--0.6160) & 0.6383 (0.5620--0.7123) & 0.6626 (0.5943--0.7323) \\
5  & 0.5826 (0.4740--0.6950) & 0.4693 (0.3851--0.5683) & 0.7504 (0.6770--0.8120) & 0.6160 (0.5214--0.7022) \\
10 & 0.5762 (0.4706--0.6919) & 0.3393 (0.2869--0.3951) & 0.7280 (0.6401--0.8051) & 0.4862 (0.4076--0.5600) \\
20 & 0.5695 (0.4668--0.6761) & 0.3231 (0.2749--0.3742) & 0.7163 (0.6216--0.7940) & 0.4706 (0.3989--0.5407) \\
40 & 0.5445 (0.4433--0.6586) & 0.3166 (0.2720--0.3659) & 0.6891 (0.5928--0.7710) & 0.4642 (0.3888--0.5320) \\
\bottomrule
\end{tabular}
}
\end{table}

At 0.5, both pathways classified every AISD slice as negative under every condition ($TP/FP/TN/FN=0/0/1002/504$). CPAISD-validation-locked thresholds yielded P1 sensitivity 0.891--0.960 with specificity 0.070--0.364, and P2 sensitivity 0.972--0.984 with specificity 0.002--0.057. Thus, neither threshold strategy established a clinically acceptable external operating point. The broader \{1,2,3,5\} sensitivity analysis produced the same qualitative pattern: P1 AP was 0.5305--0.6389 and ROC--AUC 0.6354--0.7579, whereas P2 was slightly higher only at setting 1 and lower at settings 5--40.

\subsection{Attribution--Mask Overlap Audit}
We performed a descriptive Grad-CAM--mask audit on all 242 lesion-bearing CPAISD test slices (Table~\ref{tab:gradcam_audit}). For the exact seed-7 input at each simulated setting, normalized maps were compared with the paired lesion masks using pointing-game accuracy, attribution mass within the mask, top-10\% IoU and Dice, and normalized peak-to-lesion-centroid distance. Across all five settings, P2 had lower mean attribution mass, top-10\% IoU, and top-10\% Dice than P1. P2 pointing-game accuracy exceeded P1 at settings 1 and 5 but was lower at settings 10--40, while normalized peak-to-lesion-centroid distance was larger for P2 at every setting. These descriptive patterns document pathway-dependent attribution changes but do not establish lesion removal, lesion localization, or causality. This analysis was not used for training, model selection, threshold selection, or calibration.

\begin{table}[!htbp]
\centering
\caption{Descriptive fixed-checkpoint Grad-CAM--mask audit on all 242 lesion-bearing CPAISD test slices. Each pathway used the exact seed-7 input realization at each image-domain simulated setting. Pointing-game accuracy is the fraction of slices whose maximum activation occurred inside the paired lesion mask; the remaining entries are slice means. These model-attribution measures are not segmentation or lesion-localization performance.}
\label{tab:gradcam_audit}
\scriptsize
\resizebox{\textwidth}{!}{%
\begin{tabular}{clccccc}
\toprule
Setting & Pathway & Pointing accuracy & Mask mass & Top-10\% IoU & Top-10\% Dice & Normalized peak--centroid distance \\
\midrule
1  & P1 (direct)            & 0.0372 & 0.0645 & 0.0845 & 0.1498 & 0.2025 \\
1  & P2 (denoise--classify) & 0.1364 & 0.0521 & 0.0428 & 0.0780 & 0.2687 \\
5  & P1 (direct)            & 0.1860 & 0.0815 & 0.0904 & 0.1557 & 0.2278 \\
5  & P2 (denoise--classify) & 0.2231 & 0.0510 & 0.0384 & 0.0697 & 0.2827 \\
10 & P1 (direct)            & 0.2355 & 0.0745 & 0.0752 & 0.1293 & 0.2483 \\
10 & P2 (denoise--classify) & 0.0207 & 0.0405 & 0.0050 & 0.0095 & 0.3445 \\
20 & P1 (direct)            & 0.2562 & 0.0724 & 0.0724 & 0.1228 & 0.2627 \\
20 & P2 (denoise--classify) & 0.0124 & 0.0396 & 0.0039 & 0.0074 & 0.3454 \\
40 & P1 (direct)            & 0.2810 & 0.0682 & 0.0632 & 0.1074 & 0.2687 \\
40 & P2 (denoise--classify) & 0.0207 & 0.0398 & 0.0042 & 0.0081 & 0.3443 \\
\bottomrule
\end{tabular}}
\end{table}

\subsection{Denoising Quality and Latency}
\begin{figure}[H]
    \centering
    \includegraphics[width=0.82\linewidth]{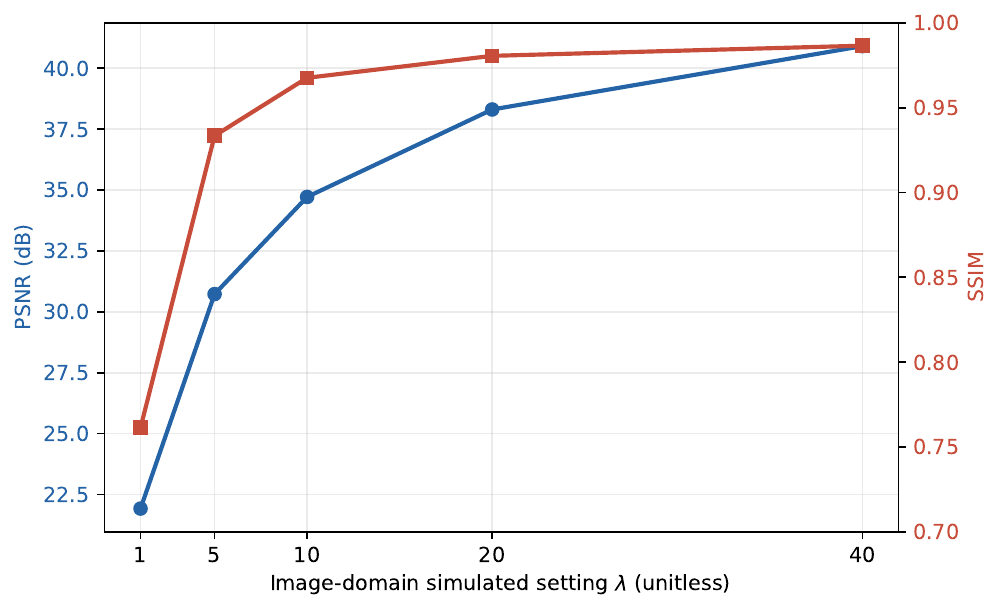}
    \caption{Fixed-checkpoint P2 output-to-reference PSNR and SSIM across unitless image-domain Poisson simulation settings. Points are means across three fixed evaluation-noise realizations; error bars are their sample SD. These values do not quantify physical acquisition dose and do not by themselves establish improvement over the unprocessed simulated input.}
    \label{fig:quality}
\end{figure}

For P2, output-to-reference fidelity increased as the unitless simulation index $\lambda$ increased (Figure~\ref{fig:quality}). Mean PSNR rose from 21.92 dB at $\lambda=1$ to 40.91 dB at $\lambda=40$, while SSIM rose from 0.761 to 0.987. These values compare P2 outputs with normalized reference images; they do not represent a physical acquisition-dose response or by themselves establish improvement over the unprocessed simulated input. Pipeline~2 required approximately 50.4--52.6 ms per slice, compared with 3.6--4.0 ms for direct classification. Thus, denoising added roughly 47--49 ms per slice in this environment.

\subsection{High-Resolution Failure-Case Audit}
\begin{figure}[H]
    \centering
    \includegraphics[width=\textwidth]{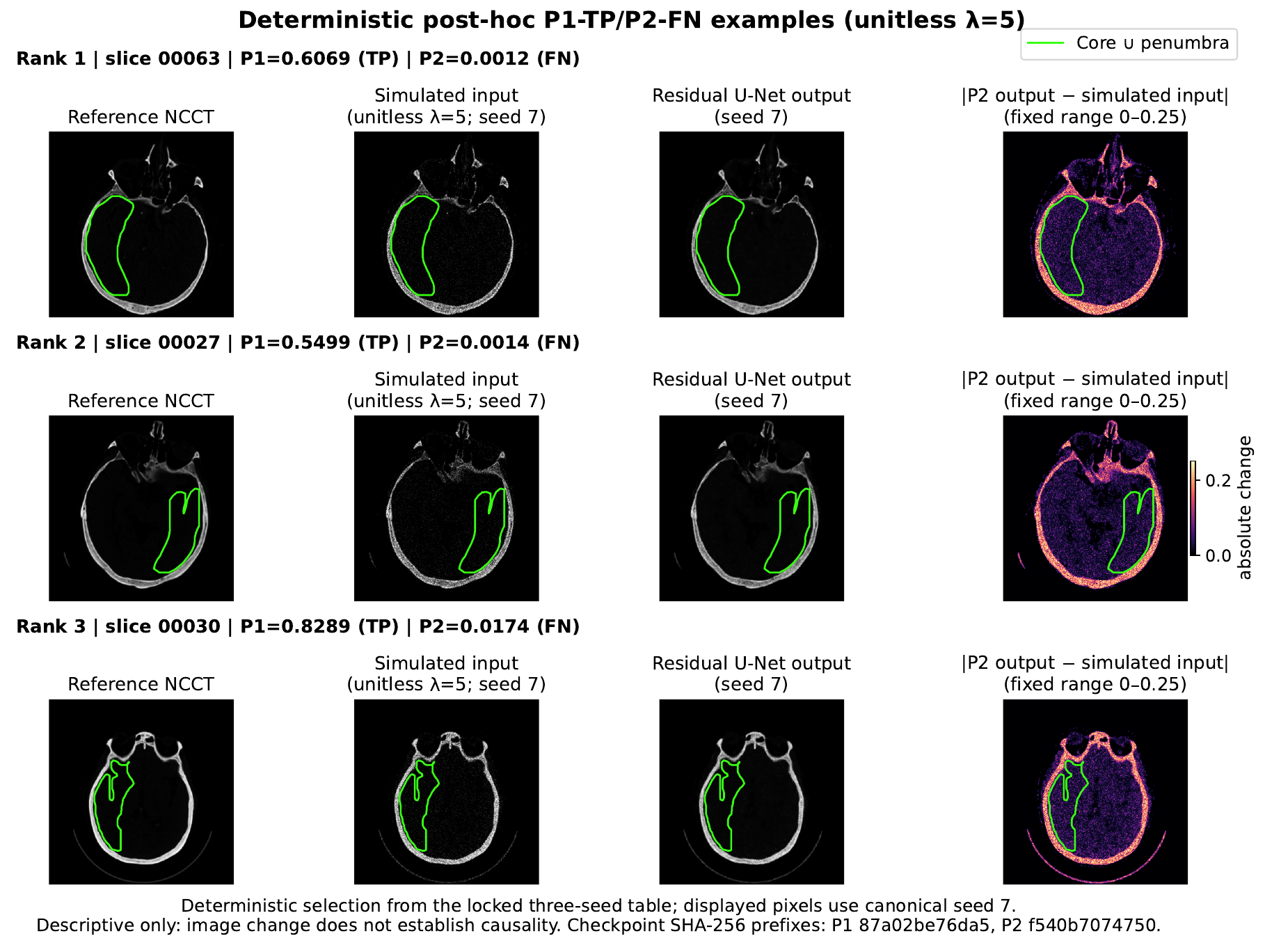}
    \caption{Deterministic post-hoc descriptive P2 failure examples at the unitless image-domain simulation setting $\lambda=5$. The rule retained the lexicographically earliest lesion-positive P1 true-positive/P2 false-negative slice from each of the three qualifying test patients after scores were averaged over seeds 7, 17, and 27 at the fixed 0.5 threshold. Panels show the reference NCCT, seed-7 simulated input, seed-7 output from the exact archived residual U-Net, and absolute input--output change; green contours denote the union of ischemic core and penumbra. The displayed seed is for visualization only; classification labels use the three-seed mean scores from the locked prediction table. This descriptive panel does not establish that any specific image change caused the score reduction.}
    \label{fig:r1_failure}
\end{figure}

At simulation setting $\lambda=5$, 31 lesion-positive slices from three CPAISD test patients were true positives with direct classification (P1) but false negatives after denoising (P2) at the fixed 0.5 threshold. To make these failures auditable without visually choosing cases, we used a deterministic post-hoc rule: the canonical test manifest was sorted by patient and slice identifier, and the lexicographically earliest qualifying slice was retained from each of the three patients. The selected P1/P2 three-seed mean scores from the locked prediction table were 0.607/0.001, 0.550/0.001, and 0.829/0.017. Figure~\ref{fig:r1_failure} shows the reference NCCT, the seed-7 simulated input, the corresponding residual U-Net output, and the absolute input--output change, with the core--penumbra union outlined in every panel. These examples document the pathway-dependent score reduction but do not establish that a particular visual change caused the classification failure.

\subsection{Qualitative Attribution Audit}
\begin{figure}[H]
    \centering
    {\small\bfseries Grad-CAM audit: exact seed-7 inputs at unitless setting $\lambda=5$\par}
    \vspace{2pt}
    {\scriptsize
    \setlength{\tabcolsep}{2pt}
    \begin{tabularx}{\textwidth}{>{\centering\arraybackslash}X >{\centering\arraybackslash}X >{\centering\arraybackslash}X >{\centering\arraybackslash}X}
    Reference NCCT + lesion mask & Poisson-perturbed NCCT (seed 7; $\lambda=5$; index 394) & Direct Grad-CAM (P1; score 0.9620) & Denoise--classify Grad-CAM (P2; score 0.0058) \\
    \end{tabularx}}
    \includegraphics[width=\textwidth,trim=0bp 144bp 0bp 25bp,clip]{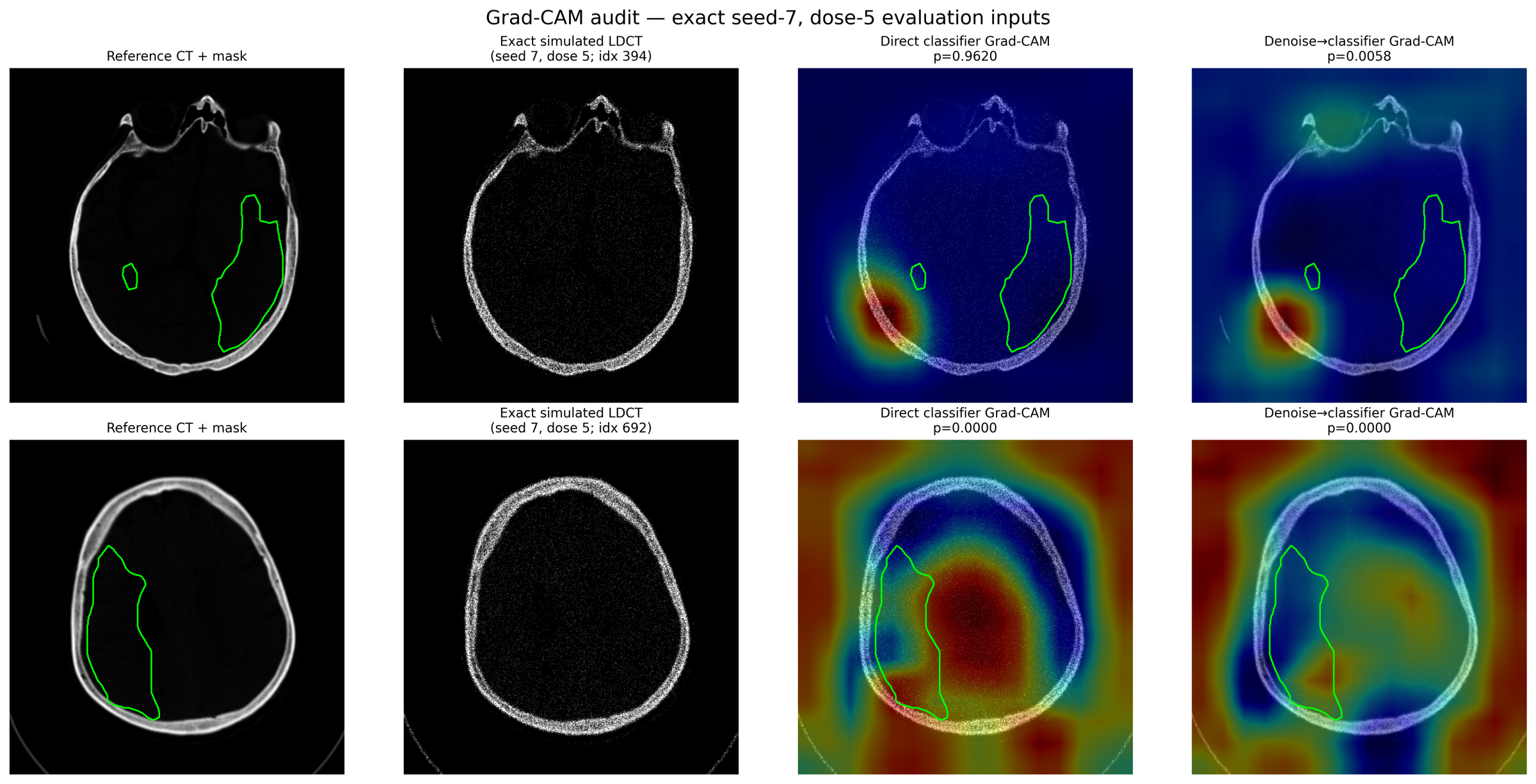}
    \vspace{2pt}
    {\scriptsize
    \setlength{\tabcolsep}{2pt}
    \begin{tabularx}{\textwidth}{>{\centering\arraybackslash}X >{\centering\arraybackslash}X >{\centering\arraybackslash}X >{\centering\arraybackslash}X}
    Reference NCCT + lesion mask & Poisson-perturbed NCCT (seed 7; $\lambda=5$; index 692) & Direct Grad-CAM (P1; score 0.0000) & Denoise--classify Grad-CAM (P2; score 0.0000) \\
    \end{tabularx}}
    \includegraphics[width=\textwidth,trim=0bp 0bp 0bp 170bp,clip]{gradcam_exact_seed7_lambda5_cases.png}
    \caption{Post-hoc illustrative fixed-checkpoint Grad-CAM audit using the exact seed-7 inputs at the unitless image-domain simulation setting $\lambda=5$; this is not acquired low-dose data. Rows show two lesion-bearing CPAISD test slices; columns show the reference NCCT with the core--penumbra union outlined in green, Poisson-perturbed input, direct-classifier Grad-CAM overlay (P1), and denoise--then--classify Grad-CAM overlay (P2). The top row is a P1 true-positive/P2 false-negative case; both pathways are false negative in the bottom row. Each heatmap is independently min--max normalized, so color represents relative within-map intensity rather than a common quantitative scale. Quantitative cohort-wide comparison is provided in Table~\ref{tab:gradcam_audit}. These maps do not validate lesion localization or establish that an attention change caused a classification error.}
    \label{fig:gradcam}
\end{figure}

To examine model attention qualitatively, we generated Grad-CAM maps from the final convolutional layer of the fixed ResNet-18 classifier for two post-hoc illustrative lesion-bearing test slices (Figure~\ref{fig:gradcam}). For each slice, the $\lambda=5$, seed-7 noisy input was reconstructed exactly using the same sorted-image index and per-slice random generator as the quantitative evaluation. We compared the direct-classifier and denoise--then--classifier maps while overlaying the reference lesion mask. The top row is a P1 true-positive/P2 false-negative example, whereas both pathways were false negative for the bottom row; displayed scores are from the illustrated seed-7 realization rather than the three-seed primary analysis. These maps make pathway-dependent attention and score changes visible, but they do not establish lesion localization, lesion preservation or removal, clinical validity, or the causal mechanism of a false-negative decision.

\subsection{Patient-Level Limitation}
All 10 CPAISD test patients had at least one lesion-bearing slice, and AISD is likewise an acute-ischemic-stroke cohort rather than a patient-negative control cohort. Mask-negative slices in either dataset are not non-stroke patients. Patient-level stroke sensitivity, specificity, predictive values, and ROC--AUC therefore cannot be estimated.

\section{Discussion}
\label{sec:discussion}
The fixed-checkpoint internal and external results separated output-to-reference fidelity from task performance. Although P2 fidelity increased with the unitless image-domain simulation index, P2 improved internal discrimination only at setting 1 and degraded it at higher settings. On independent AISD DICOM/HU inputs, the P2 advantage at setting 1 was inconclusive and P2 was significantly worse at settings 5--40. These observations were consistent with pathway-dependent input distribution shift but could not establish whether task-aware joint training would help.

The completed de novo ablation directly tested that question. DNC had the highest across-setting mean AP point estimate (0.8606), JDC-0 was lower (0.8399), and JDC-B recovered most of that point-estimate difference (0.8557). However, the hierarchical 95\% intervals for JDC-0 minus DNC, JDC-B minus JDC-0, and JDC-B minus DNC all included zero. We therefore found no statistically supported classification benefit from ordinary joint training, from this specific boundary term, or from the full boundary-aware system over direct classification. The result is informative despite being null: improving or constraining the reconstruction objective did not reliably improve ranking in this implementation.

At $\lambda=5$, the fixed-checkpoint case audit showed that 31 lesion-positive slices spanning three patients crossed from P1 true-positive to P2 false-negative status (Figure~\ref{fig:r1_failure}). The absolute-change maps identify where the residual U-Net altered each displayed image, but they cannot distinguish lesion attenuation from broader intensity shift or lack of score alignment/calibration relative to the CPAISD audit endpoint as the cause. The new ablation also did not isolate a single mechanism. Oversmoothing, objective mismatch, optimization, calibration, and distribution shift therefore remain hypotheses rather than causal conclusions; these findings should not be generalized to denoising as a class of methods.

The external stress test also tempers the apparent internal ranking performance. Unperturbed AISD reference NCCT was near chance, and the higher ranking observed after some synthetic perturbations is counterintuitive. It must not be interpreted as evidence that radiation-dose reduction improves diagnosis. Instead, it indicates that the frozen checkpoint is highly sensitive to intensity and preprocessing distributions. AISD nevertheless adds useful independent evidence: the adverse P2-versus-P1 effect at settings 5--40 reproduced across 52 external patients, with paired patient-cluster intervals excluding zero.

Calibration was a separate and severe limitation. A nominal 0.5 cutoff produced few or no positive predictions internally and all-negative predictions externally. Thresholds locked on CPAISD validation restored high sensitivity only at the cost of very low specificity, especially for P2. Strong ranking metrics at selected conditions therefore cannot be equated with a clinically usable decision rule. Independent calibration and prospective evaluation would be required before any operating point could be proposed.

Table~\ref{tab:p2_failure_evidence} distinguishes the observed decision failure from candidate upstream explanations. The available evidence identifies severe score compression and lack of alignment/calibration relative to the CPAISD audit endpoint at the prespecified threshold, but it does not isolate a unique historical training or image-transformation mechanism.

\begin{table}[H]
\centering
\caption{Evidence--hypothesis reconciliation for the historical Pipeline~2 sensitivity failure. ``Observed'' denotes a directly measured result; candidate mechanisms are retained only to the extent supported by the listed comparison.}
\label{tab:p2_failure_evidence}
\scriptsize
\begin{tabularx}{\textwidth}{>{\raggedright\arraybackslash}p{0.17\textwidth} >{\raggedright\arraybackslash}X >{\raggedright\arraybackslash}p{0.30\textwidth}}
\toprule
Issue & Verified evidence & Supported interpretation and limit \\
\midrule
Decision threshold and calibration & At 0.5, P2 yielded $TP/FP/TN/FN=0/0/567/242$ at every fixed-checkpoint setting; its positive-slice median scores were 0.001029 or lower after setting 1, and its largest score in any condition was 0.189031. Validation-locked thresholds restored 100\% test sensitivity but only 0.5--1.1\% specificity; AP and ROC--AUC nevertheless retained ranking information in several settings. & Observed score compression and lack of alignment/calibration relative to the CPAISD audit endpoint explain the zero sensitivity at 0.5. They do not identify the upstream cause or provide a clinically usable operating point. \\
Class imbalance & The patient-disjoint training partition contained 1,719/6,657 lesion-positive/mask-negative slices. The de novo DNC, JDC-0, and JDC-B models used this same partition and unweighted BCE, yet none reproduced the all-negative collapse; their fixed-0.5 sensitivity ranges were 0.664--0.687, 0.678--0.716, and 0.632--0.674, respectively. & The observed imbalance is not sufficient by itself to explain the historical collapse. A class-reweighting or sampling ablation was not performed, so a contributory interaction cannot be excluded. \\
Loss and task-aware objective & Historical loss coefficients and stopping provenance are unavailable. In the locked de novo experiment, JDC-0 minus DNC was $-0.0207$ (95\% CI $-0.0625$ to $0.0188$), and JDC-B minus JDC-0 was $0.0158$ ($-0.0188$ to $0.0542$). & Neither ordinary joint training nor the tested boundary term showed a statistically supported across-setting mean AP benefit. This does not compare alternative class-weighted or focal losses or recover the historical cause. \\
Oversmoothing or lesion-information loss & Figure~\ref{fig:r1_failure} documents image changes in P1-true-positive/P2-false-negative cases. Across all five settings, P2 had lower Grad-CAM mask mass, top-10\% IoU, and top-10\% Dice and larger normalized peak--centroid distance than P1 (Table~\ref{tab:gradcam_audit}). & The results document pathway-dependent image and attribution changes. Without a controlled smoothing intervention, they do not establish lesion removal, oversmoothing, localization performance, or causality. \\
Input-distribution sensitivity & On the 52-patient AISD stress test, P2 was worse than P1 at settings 5--40; paired AP differences ranged from $-0.2464$ to $-0.1132$ and paired ROC--AUC differences from $-0.2457$ to $-0.1344$, with corresponding 95\% intervals excluding zero. & The frozen system is sensitive to the input distribution, but this comparison does not separate distribution shift from calibration, reconstruction, or other historical training factors. \\
\bottomrule
\end{tabularx}
\end{table}

The quantitative Grad-CAM audit replaces purely qualitative saliency interpretation with a mask-based descriptive check, but the modest overlap does not establish lesion localization. Similarly, the P1-versus-P2 comparison is an inference-stage component test, not an architecture benchmark or causal training ablation.

\subsection{Scope and Limitations}
CPAISD contains 112 hyperacute-ischemic-stroke patients and only 10 test patients; AISD contributes 52 external test patients but uses related, nonidentical infarct labels. Neither cohort provides suitable non-stroke control patients for patient-level diagnosis. Patient-cluster and hierarchical bootstrapping respect dependence but cannot overcome the small number of independent test patients, scanner mix, annotation shift, or case-selection limitations. The image-domain Poisson perturbation omits projection statistics, scanner-specific correlations, reconstruction kernels, beam hardening, motion, ring artifacts, and cone-beam geometry. Thus, AISD is an external conventional-NCCT stress test, not real acquired low-dose, portable-CT, or CBCT validation. Five paired seeds quantify training variability only within the tested implementation. JDC-B also requires privileged lesion masks during training, increasing annotation requirements. Although the new experiment was locked before its one-time test evaluation, earlier fixed-checkpoint test findings were already known; it was therefore not fully blinded. The historical checkpoint training logs remain incomplete, and no clinically validated operating point has been established. Prospective work should use a larger patient-balanced multicenter cohort, independent calibration, reader and lesion-focused evaluation, and genuinely acquired low-dose or portable imaging.

\section{Conclusion} \label{sec:conclusion}
In the fixed-checkpoint audit, P2 output-to-reference fidelity increased as the unitless simulation index increased, but residual U-Net preprocessing did not consistently improve ischemic-lesion slice ranking, and the external AISD stress test exposed limited transportability. The five-seed de novo experiment likewise showed no statistically supported across-setting mean AP benefit for joint denoising--classification over direct classification; adding the tested lesion-boundary objective recovered part of the point-estimate loss but did not yield a significant incremental or full-system advantage. Image fidelity, ranking, calibration, and clinical utility are therefore distinct properties that require separate evaluation. The evidence supports a controlled image-domain Poisson-perturbation robustness and task-aware ablation study, not patient-level diagnosis or validation of acquired low-dose, portable, or cone-beam CT. Future work should independently replicate the fully logged experiment in larger patient-balanced multicenter cohorts with non-stroke controls, compare additional task-aware reconstruction objectives, perform prospective calibration and reader studies, and evaluate genuinely acquired low-dose, portable-CT, and cone-beam-CT data.

\section{Data and Code Availability}
The canonical CPAISD source dataset is available from Zenodo at \url{https://doi.org/10.5281/zenodo.10892316}, with its peer-reviewed descriptor \cite{umerenkov2025cpaisd}. The derived Kaggle packaging used here is available at \url{https://www.kaggle.com/datasets/coolrhea/ldct-classification}. AISD, its official 52-case test identifiers, original DICOM downloads, label definitions, and use terms are provided by the dataset authors at \url{https://github.com/GriffinLiang/AISD} \cite{liang2021symmetry}. The versioned evaluation notebook at \url{https://www.kaggle.com/code/coolrhea/ldct-final-evaluation} provides the fixed-checkpoint predictions, bootstrap tables, threshold audits, figures, and reproducibility manifests. The completed de novo experiment retains all 15 run logs, checkpoint hashes, locked manifests, final predictions, and machine-readable summaries. Its verified per-seed condition and reconstruction tables are provided as ancillary files with this preprint as Supplementary Data~S1 and S2, respectively. The four-decimal cohort-level Grad-CAM summary underlying Table~\ref{tab:gradcam_audit} is provided as Supplementary Data~S3 (\nolinkurl{gradcam_mask_summary_4dp.csv}), and the fixed-checkpoint class-conditional score summaries are provided as Supplementary Data~S4 (\nolinkurl{fixed_checkpoint_score_distributions.csv}). A compact, checksum-indexed provenance package for all 15 de novo runs and the sealed final evaluation is available from the corresponding author; large model checkpoints remain archived with their SHA-256 identities and are also available from the corresponding author.

\section{Ethics Statement}
This secondary analysis used publicly available, anonymized datasets and involved no new participant recruitment or data collection by the authors. Ethics review and consent procedures for the source cohorts are described in their original publications and dataset documentation.

\section{Author Contributions}
RhG: Conceptualization, Methodology, Software, Formal analysis, Investigation, Visualization, Writing--original draft, and Writing--review and editing. RoG: Conceptualization, Data curation, Methodology, Supervision, and Writing--review and editing. EL: Writing--review and editing. All authors reviewed and approved the manuscript.

\section{Funding}
The authors declare that no financial support was received for the research, authorship, or publication of this article.

\section{Conflict of Interest}
The authors declare that the research was conducted in the absence of any commercial or financial relationships that could be construed as a potential conflict of interest.

\section{Acknowledgments}
The authors thank Dr. Mingji Lou for mentorship and guidance. As disclosed in the Reproducibility subsection, generative AI technology (OpenAI GPT-5, accessed through Codex) assisted with code review, reproducibility checks, and language and LaTeX editing; all outputs were executed or checked by the authors, who take responsibility for the work.

\FloatBarrier
\bibliographystyle{unsrtnat}
\bibliography{references}
\clearpage
\section*{Figure captions and figures}

\end{document}